\pdfoutput=1

\documentclass[11pt]{article}

\usepackage[final]{acl}

\usepackage{times}
\usepackage{latexsym}

\usepackage[T1]{fontenc}

\usepackage[utf8]{inputenc}

\usepackage{microtype}

\usepackage{inconsolata}

\usepackage{graphicx}

\usepackage{hyperref}
\usepackage{url}            
\usepackage{booktabs}       
\usepackage{amsfonts}       
\usepackage{nicefrac}       
\usepackage{xcolor}         
\usepackage{float}
\usepackage{multirow}
\usepackage{amsthm}
\usepackage{xcolor}
\usepackage{comment}
\usepackage{mathtools}

\usepackage{todonotes}
\usepackage{colortbl}
\usepackage{enumitem}
\usepackage{bm} 
\usepackage{hwemoji}
\usepackage{pifont}

\title{How Well Can a Long Sequence Model Model Long Sequences? Comparing Architectural Inductive Biases on Long-Context Abilities}

\author{%
    Jerry Huang \\ 
    Mila - Quebec AI Institute \& Universit\'{e} de Montr\'{e}al \\
    \href{mailto:jerry.huang@mila.quebec}{\texttt{jerry.huang@mila.quebec}}
}

\begin{document}

\maketitle

\begin{abstract}
Long sequences occur in abundance within real-world scenarios, hence properly modelling them opens numerous down-stream use-cases. Deep neural networks, however, have often struggled with these for a variety of reasons. Recent advances, both in system engineering as well as model design, have enabled the scaling up of model that are purported to support extended context length. In particular, the state-space and linear recurrent neural network families of models hypothetically can entend to infinite sequence lenth. However, is this too good to be true? We conduct a targeted evaluation, where we show that while such claims may have theoretical soundness under particular conditions, these may break down in practical settings where limitations exist. In particular, we observe that these new-age sequence models suffer similarily as attention-based models when it comes to long-contexts, highlighting the need to further study such paradigms and why they seemingly fail to behave as expected.
\end{abstract}

\section{Introduction}

Advances in AI system engineering~\citep{dao2022flashattention, dao2024flashattention, deepspeed} and model design~\citep{linearattention, jiang2023mistral, jamba} have opened language models to the broader public for a diverse set of purposes and use cases. However, Transformer-based architechtures~\citep{vaswani2017attention} remain bounded in terms of their context windows, as they require fixed-length positional embedding representations~\citep{press2022train, su2023roformer, peng2024yarn} which cannot be modified a posteriori. With this glaring limitation, linear sequence models~\citep{gu2022efficiently, gu2024mamba, lru, qin2023hierarchically, peng-etal-2023-rwkv, de2024griffin, dao2024transformers} have emerged as an alternative that present a seeming ability to extend to infinite-length contexts in theory while retaining all the original benefits of the Transformer related to training-based parallization.

However, despite the temptation to assert linear sequence models as superior, properly testing for information retention from long-context tasks remains callenging. Although some work has attempted to evaluate this ability through long contexts~\citep{shaham-etal-2022-scrolls, pang-etal-2022-quality, dong-etal-2024-bamboo-comprehensive, bai2023longbench, li2023loogle, han2024lminfinite}, whether or not they truly require the use of long-contexts is uncertain and ascertaining long-context abilities from these tasks is difficult. This has prompted the use of more synthetic tasks~\citep{hsieh2024ruler}, such as needle-in-a-haystack (NIAH)~\citep{needle} and passkey retreival~\citep{mohtashami2023landmark}, to better control and evaluate the context sizes of models.

Nevertheless, an outstanding question remains whether or not long-context models can effectively model long contexts. While some works~\citep{gu2024mamba, fu2023hungry, hyena, peng2024yarn, qwen2} purport to be able to extrapolate towards sequences of long length (100k tokens+), further investigation has suggested differently. For example, \citet{hsieh2024ruler} claim modern LLMs significantly over-state true context windows on a number of synthetic tasks. Meanwhile \citet{han2024lminfinite} observe models to perform reasonably well on synthetic tasks, but struggle on real-world tasks, as do \citet{li2023loogle}. Hence despite a consistent trend in models behaving underwhelmingly, it remains to be understood why this occurs. Yet one interesting question is whether or not linear sequence models are in fact more suited for these compared to Transformer-based ones, as has been claimed repeatedly.

To this end, we further analyze the behaviour of sequence models to observe how differently they behave compared to Transformer-based ones. We perform a more extensive study into each type of model, as well as a mixture of both, to better investigate how they perform in principle and how they change in behaviour when extending to longer and longer sequences. On both synthetic and realistic data, we conduct a thorough study and observe:
\begin{itemize}[itemsep=0ex, leftmargin=0.5cm]
    \item All models, whether they use pure sequence layers, attention or a mix, struggle with extrapolating beyond their training context length.
    \item The abiliy to extrapolate can vary signficantly based on the format of the sequence even if the task remains constant. However models consistently struggle more with information placed in the middle of long contexts.
\end{itemize}
These results highlight that long sequence models suffer from significant limitations despite their theoretical soundness, highlighting a need to better understand this striking dissonance between expectation and observation and how to amend it for better long-context understanding and reasoning.

\section{Related Work}

\paragraph{Efficient Long-Context Models.} Due to the computational bottleneck of attention~\citep{bahdanau2016neural} relative to sequence length, significant modifications have been made to overcome this limitation of the Transformer~\citep{child2019generating, linearattention, su2023roformer} yet they remain theoretically bounded in terms of its context length. Alternatively, sequence models~\citep{rumelhart1987parallel, jordan1986serial, lstm, cho-etal-2014-learning} originally faced significant issues that limited their application but recent modifications~\citep{gu2020hippo, gu2021combining} have led to the prominence of linear sequence models which are significantly more compute-effective than Transformer-based architechtures.

\paragraph{On the Limits of Long Sequence Models.} Due to their more intuitive and interpretable architechture, long/linear sequence models remain easier to analyze when placed in comparision to Transformers. As such, their limitations also become easier to discover and analyze. \citet{vardasbi-etal-2023-state} first show that SSMs struggle at sequence-to-sequence tasks due to to the use of a fixed-size hidden representation which compresses the entire prior context, making it difficult to extract information from the past, fact further substantiated by \citet{jelassi2024repeat}. \citet{park2024can} additionally demonstrate that these models have difficulty with more complex in-context learning tasks, while \citet{merrill2024the} show them to possess similar limitations in terms of representational power as Transformers~\citep{merrill2023parallelism}. \citet{waleffe2024empirical} finally make a comparision between Mamba, Transformers as well as a hybrid and observe hybrid models to perform better on long-context tasks, while Mamba2 often trails behind Transformers. These observations thus beg a question: can long sequence models really model long sequences? Given the hints that long sequence models may not always be as they seem, a more formal investigation is necessary. We distinguish ourselves by conducting a more controlled but intricate study which aims to uncover why some of the prior results might occur, which we discuss in the work that follows.

\section{Background}

\paragraph{Attention and Long Sequences.} Self-attention as used in Transformers is powerful but costly. When provided an embedded text representation as a sequence of tokens $\bm{x}\in\mathbb{R}^{L \times d}$, each Transformer layer in the network applies a function
\begin{equation}
    T_\ell(\bm{x}) = \text{FF}_\ell(A_\ell(\bm{x}) + \bm{x}) + A_\ell(\bm{x})
    \label{eq:transformer_layer}
\end{equation}
where $A_\ell$ is the self-attention mechanism of the $\ell$-th layer and $\text{FF}_\ell$ is the following feed-forward network\footnote{Excludes normalization operations.}. Self-attention computes, for every position, a weighted average of the feature representations of
all other positions with a weight proportional to a similarity score between the representations.
\begin{equation}\label{eq:attention}
    \begin{split}
    &\bm{Q}_\ell = \bm{x}\bm{W}_\ell^{\bm{Q}} \quad \bm{K}_\ell = \bm{x}\bm{W}_\ell^{\bm{K}} \quad \bm{V}_\ell = \bm{x}\bm{W}_\ell^{\bm{V}} \\
    &A_\ell(\bm{x}) = \bm{V}_\ell' = \text{softmax}\big({\bm{Q}_\ell\bm{K}_\ell^T}/{\sqrt{d}}\big)\bm{V}_\ell
    \end{split}
\end{equation}
As the softmax operation operates in $O(L^2)$ time when applied naively, this limits the ability to process long-sequences.
\paragraph{Transformers to Sequence Models.} The long-sequence limitations of Transformers necessitates the need for alternatives in such settings, which have currently appeared under the form of state-of-the-art sequence models. An initial proposal borrowed from control theory, namely the notion of state-space models (SSMs). These model a dynamical system, traditionally mapping a 1-D continuous input signal $x(t)\in\mathbb{R}$ to an $n$-dimensional hidden state $h(t)\in\mathbb{R}^n$ that is projected back to a 1-D output $y(t)\in\mathbb{R}$ using:
\begin{equation}\label{eq:ssm_cont}
    \begin{cases}
        h'(t) &= {\bm{A}}h(t) + {\bm{B}}x(t) \\
        y(t) &= {\bm{C}}h(t) + {\bm{D}}x(t)
    \end{cases}
\end{equation}
where $\bm{A}$, $\bm{B}$, $\bm{C}$ and $\bm{D}$ are all trainable parameters. \citet{gu2021combining} use this paradigm to define a recurrent model to work on discrete signals, in which case the input can be regarded as discretized data sampled from a continuous signal with a step size $\Delta$, for which the corresponding SSM is defined by:
\begin{equation}\label{eq:ssm}
    \begin{split}
        h_{t} &= \overline{\bm{A}}h_{t-1} + \overline{\bm{B}}x_{t} \quad y_{t} = \overline{\bm{C}}h_{t} + \overline{\bm{D}}x_{t} \\
        \overline{\bm{A}} &= \frac{\big(I+{\Delta}\bm{A}/{2})}{\big(I-{\Delta}\bm{A}/{2}\big)} \quad \overline{\bm{B}} = \frac{\Delta\bm{B}}{\big(I-{\Delta}\bm{A}/{2}\big)} 
    \end{split}
\end{equation}
and $\overline{\bm{C}} = \bm{C}$ (They set $\overline{\bm{D}} = 0$ due to being equivalent to a residual connection.) Thus the output $\bm{y}$ given an input $\bm{x}$ is
\begin{equation}\label{eq:ssm_kernel}
    \begin{split}
        \overline{\bm{K}} &= (\overline{\bm{CB}}, \overline{\bm{CAB}}, \dots, \overline{\bm{CA}}^{L-1}\overline{\bm{B}}) \\
        y_t &= \sum_{j=0}^{L-1} \overline{\bm{CA}}^{j}\overline{\bm{B}} x_{L-j}= \overline{\bm{K}} * \bm{x}
    \end{split}
\end{equation}
where $\overline{\bm{K}}$ is the SSM kernel. As $\bm{y}$ can be computed in $O(L\log L)$ with a Fast Fourier Transform~\citep{algorithms}, the entire output can be computed in tandem based on the input, given the matrices that parametrize the system. \citet{gu2021combining} use this to overcome issues of parallelization and vanishing gradients~\citep{bengio1994learning, hochreiter2001gradient, pascanu2013difficulty} observed by prior recurrent models by
\begin{enumerate}[label=(\arabic*), ref=\arabic*]
    \item Removing non-linearities in the recurrence, enabling the efficient pre-computation of $\overline{\bm{K}}$.
    \item Using a special matrix parameterization~\citep{gu2020hippo} for $\bm{A}$  to memorize the input and eliminate exponential gradient scaling.
\end{enumerate}
This has sparked a new wave of recurrent models to compete with Transformers~\citep{lru, qin2023hierarchically, de2024griffin, beck2024xlstm}, with the added benefit of theoretically having longer context sizes that scale more efficiently.

\section{Experiments and Results}

\paragraph{Datasets.} We conduct an initial evaluation using \textsc{Ruler}~\citep{hsieh2024ruler}, a set of synthetic benchmarks that test long-context information retention, before conductin a more fine-grained evaluation on a general needle-in-the-haystack task. We use this benchmark as for more granular control over the exact information that must be retained. Results are measured in terms of accuracy based on exact matching of predicted tokens.

\paragraph{Baselines.} Our main objective is to compare how long-sequence models fare on long context tasks. To this end, we compare models with the same number of parameters that are evenly trained on the same data. Hence we first use Mamba2~\citep{dao2024transformers} as well as a Transformer variant (Transformer++) as well as a hybrid Mamba2Attn, each with 2.7 billion parameters. We further add Sheared-LLaMA~\citep{xia2024sheared} and RecurrentGemma~\citep{botev2024recurrentgemma} baselines (with and without intruction-tuning) as same-sized baselines trained under different conditions. We finally add a 3 billion RWKV~\citep{peng-etal-2023-rwkv} variant as another sequence model baseline.

\paragraph{Results.} We present initial results on the base set of \textsc{Ruler} tasks (as defined by its original authors) in \autoref{tab:ruler_results}. Results presented are averaged across individual tasks within the benchmark, which are described in futher detail in \autoref{app:ruler}. However, we present two additional ablation studies. In the first, we use a single needle hidden within a large haystack; however, we modify its relative position within the context. The goal of this ablation, presented in \autoref{tab:niah_position} and \ref{tab:niah_position_2}, is to observe how the use of a unified hidden state rather than attention can affect the ability to retain information throughout a long sequence. The second (\autoref{tab:niah_ablation}) further tests how this information retention may change when the content that is being memorized changes (e.g. numbers versus UUIDs within a haystack of repeated sentences or essays). In all tables, we abbreviate model names using titles noted in \autoref{app:implementation_details}.

\begin{table}[ht!]
    \resizebox{\linewidth}{!}{
        \centering
        \begin{tabular}{l|ccccc|c}
        \toprule
        Length & \texttt{1K} & \texttt{2K} & \texttt{4K} & \texttt{8K} & \texttt{16K} & Average \\
        \midrule
        \texttt{Mamba2} & 38.52 & \underline{32.91} & 12.98 & 6.51 & 0.1 & 18.2 \\
        \texttt{M2A} & 39.14 & \underline{30.43} & 12.89 & 7.8 & 3.49 & 18.75 \\
        \texttt{TPP} & 46.61 & \underline{36.74} & 0.31 & 0.06 & 0.03 & 16.75 \\
        \midrule
        \texttt{RG} & 78.82 & \underline{71.72} & 22.45 & 11.21 & 6.29 & 38.1 \\
        \texttt{SL} & 84.38 & 69.89 & \underline{58.37} & 0.0 & 0.0 & 42.53 \\
        \texttt{RWKV} & 68.09 & 55.27 & \underline{37.47} & 23.73 & 13.81 & 39.67 \\
        \midrule
        \texttt{RG-IT} & 85.64 & \underline{79.45} & 44.33 & 24.19 & 14.18 & 49.56 \\
        \texttt{SL-IT} & 86.22 & 77.54 & \underline{74.25} & 0.0 & 0.0 & 47.6 \\
        \bottomrule
        \end{tabular}
    }
    \caption{Results on \textsc{Ruler}. Accuracy is aggregated across several tasks for each model and context length. Context length for which each model was trained is underlined. Best performing models are bolded.}
    \label{tab:ruler_results}
\end{table}

\begin{table}[ht!]
    \resizebox{\linewidth}{!}{
        \centering
        \begin{tabular}{l|ccccccc|c}
        \toprule
        Position & \texttt{0} & \texttt{20} & \texttt{40} & \texttt{50} & \texttt{60} & \texttt{80} & \texttt{100} & \texttt{Avg} \\
        \midrule
        \texttt{Mamba2} & 59.07 & 31.47 & 33.07 & 39.07 & 40.0 & 31.33 & 66.0 & 42.63 \\
        \texttt{M2A} & 40.27 & 36.53 & 30.27 & 29.33 & 29.33 & 35.07 & 37.2 & 35.26 \\
        \texttt{TPP} & 53.33 & 33.47 & 22.8 & 26.27 & 31.33 & 35.07 & 55.73 & 35.64 \\
        \midrule
        \texttt{RG} & 100.0 & 100.0 & 100.0 & 100.0 & 100.0 & 100.0 & 99.47 & 99.92 \\
        \texttt{SL} & 99.6 & 99.6 & 100.0 & 100.0 & 100.0 & 100.0 & 100.0 & 99.89 \\
        \texttt{RWKV} & 82.4 & 100.0 & 100.0 & 80.27 & 100.0 & 100.0 & 100.0 & 94.67 \\
        \midrule
        \texttt{RG-IT} & 100.0 & 100.0 & 100.0 & 100.0 & 100.0 & 100.0 & 100.0 & 100.0 \\
        \texttt{SL-IT} & 98.27 & 99.6 & 100.0 & 100.0 & 100.0 & 100.0 & 99.73 & 99.66 \\
        \bottomrule
        \end{tabular}
    }
    \caption{Results on needle-in-a-haystack task where the position of a single needle is at a fixed depth within the haystack. Context length is set to the maximum on which the models were trained.}
    \label{tab:niah_position}
\end{table}
\begin{table}[ht!]
    \resizebox{\linewidth}{!}{
        \centering
        \begin{tabular}{l|ccccccc|c}
        \toprule
        Position & \texttt{0} & \texttt{20} & \texttt{40} & \texttt{50} & \texttt{60} & \texttt{80} & \texttt{100} & \texttt{Avg} \\
        \midrule
        \texttt{Mamba2} & 26.8 & 19.6 & 17.73 & 18.93 & 18.93 & 20.13 & 21.87 & 21.03 \\
        \texttt{M2A} & 38.8 & 26.27 & 18.93 & 28.8 & 10.13 & 21.6 & 66.67 & 27.07 \\
        \texttt{TPP} & 0.0 & 0.0 & 0.0 & 0.0 & 0.0 & 0.0 & 0.0 & 0.0 \\
        \midrule
        \texttt{RG} & 0.0 & 0.0 & 0.0 & 99.87 & 100.0 & 100.0 & 96.27 & 56.59 \\
        \texttt{SL} & 0.0 & 0.0 & 0.0 & 0.0 & 0.0 & 0.0 & 0.0 & 0.0 \\
        \texttt{RWKV} & 33.47 & 99.6 & 100.0 & 36.53 & 100.0 & 100.0 & 100.0 & 81.37 \\
        \midrule
        \texttt{RG-IT} & 0.0 & 0.0 & 0.0 & 100.0 & 99.6 & 100.0 & 99.73 & 57.05 \\
        \texttt{SL-IT} & 0.0 & 0.0 & 0.0 & 0.0 & 0.0 & 0.0 & 0.0 & 0.0 \\
        \bottomrule
        \end{tabular}
    }
    \caption{Same results as above with context length set to twice the maximum training length.}
    \label{tab:niah_position_2}
\end{table}


\begin{table}[ht!]
    \resizebox{\linewidth}{!}{
        \centering
        \begin{tabular}{l|c|ccc|ccc|ccc}
        \toprule
        \multirow{2}{*}{Model} & Context & \multicolumn{3}{c|}{\tt Essay-Word-Num} & \multicolumn{3}{c|}{\tt Essay-Word-UUID} & \multicolumn{3}{c}{\tt Repeat-Word-Num} \\
        & Length & \texttt{0} & \texttt{50} & \texttt{100} & \texttt{0} & \texttt{50} & \texttt{100}& \texttt{0} & \texttt{50} & \texttt{100} \\
        \midrule
        \multirow{3}{*}{\texttt{Mamba2}} & 1024 & 86.0 & 73.6 & 82.0 & 78.0 & 70.8 & 80.8 & 77.6 & 70.4 & 55.2 \\ & 2048 & 45.6 & 20.8 & 65.2 & 49.6 & 20.4 & 66.0 & 82.0 & 76.0 & 66.8 \\ & 4096 & 0.0 & 0.0 & 0.0 & 0.0 & 0.0 & 0.0 & 80.4 & 56.8 & 65.6 \\
        \midrule
        \multirow{3}{*}{\texttt{M2A}} & 1024 & 37.2 & 28.0 & 48.0 & 39.2 & 26.8 & 48.0 & 47.2 & 44.4 & 70.0 \\ & 2048 & 41.6 & 27.6 & 39.6 & 42.4 & 28.4 & 30.8 & 36.8 & 32.0 & 63.2 \\ & 4096 & 29.2 & 25.6 & 59.2 & 27.6 & 28.0 & 58.0 & 59.6 & 32.8 & 82.8 \\
        \midrule
        \multirow{3}{*}{\texttt{TPP}} & 1024 & 52.0 & 36.0 & 47.6 & 58.8 & 34.4 & 50.4 & 81.6 & 33.2 & 58.4 \\ & 2048 & 51.6 & 29.6 & 62.4 & 44.8 & 36.0 & 55.6 & 63.6 & 13.2 & 49.2 \\ & 4096 & 0.0 & 0.0 & 0.0 & 0.0 & 0.0 & 0.0 & 0.0 & 0.0 & 0.0 \\
        \bottomrule
        \end{tabular}
    }
    \caption{Results on needle-in-a-haystack task where the position of a single needle is placed at the beginning, end or middle of the haystack while the types of each component varies. Context length is set to the maximum on which the models were trained.}
    \label{tab:niah_ablation}
\end{table}

\section{Discussion}

\paragraph{All models have limits.} Our first observation is that regardless of the model, performance drops steeply upon testing with sequences that are longer than what the model was initially trained on. This is made clear in \autoref{tab:ruler_results}, where the performance decline is greatest once the evaluated sequences are longer than the training context (with the mild exception of RWKV which demonstrates approximately linear degredation as the sequences progressively double in length). However, an important observation is that linear sequence models do appear to extrapolate slightly better than pure-attention models, whose performance drop to near 0 performance upon the increase, as these models do show non-trivial accuracy even when evaluated on the longer sequences. This distinction is less clear when comparing between pure linear sequence models and hybrid models which alternate between sequence-model layers and attention layers, as there is no explicit pattern as to when one class will perform better on one length or another.

\paragraph{Being lost in the middle is a common event.} Being lost in the middle, whereby models have difficulty recalling relevant information located positionally in the middle of long contexts~\citep{lostmiddle}, has been observed as a common limitation among attention-based models. In \autoref{tab:niah_position}, this appears to be a common feature among all models we test, since all classes of models see increasing drops in performance as the information is more closely located at the center of the sequence. This suggests that despite their long-context modeling ability, recurrent models cannot effectively reason over their entire context window when prompted. However, when extending beyond the length of the training context (\autoref{tab:niah_position_2}), there is less consistency in the pattern, but models generally remain more capable when information is close to either end of the sequence. Moreover, while Mamba models still appear lost-in-the-middle, other recurrent models such as RecurrentGemma and RWKV have no clear depth-to-performance trends, further bringing into question their general long-context modeling abilities and how they function.


\paragraph{Extrapolation can inconsistent.} Furthermore, extrapolation can be inconsistent based on characteristics of the model as well as the data. In \autoref{tab:niah_ablation}, we can first note that, depending on the data format of the haystack, key, and value to be retrieved, the performance of each model varies significantly even with the same task template, context length and needle position. Furthermore, extrapolation varies based on the model as these characteristics change. For example, pure sequence layers (\texttt{Mamba2}) appear to only extrapolate when the haystack is a repeated sequence and the retrived value is a number related to a key word. Upon changing the haystack to be essays, extrapolation craters, and the model fails. An equally trained hybrid model (\texttt{M2A}) can meanwhile always extrapolate to some degree, but performance on sequences up to the training context length appears to compare much worse. Pure attention (\texttt{TPP}) meanwhile performs favorably only when evaluating on the extact training context length under specific data formats, but otherwise underwhelms.

\section{Conclusion}

In this work, we conduct a comprehensive comparision between the long-sequence models and attention-based language models, showing that long-context abilities of such sequence models may hold from a theoretical perspective, they empirically still struggle in comparison to models that make no guarantees. This highlights the need to improve long sequence reasoning abilties not only for Transformer-based LLMs, but also SSMs and new classes of RNNs, which hopefully can serve as motivation to further analyze this topic.

\section{Limitations}

We limit ourself to a model size in which it is easy to compare models of various paradigms. As such, some perhaps more powerful models are not explored as the analysis between such models can become difficult due to multiple additional changing variables that can perhaps lead to incorrect or undersupported claims.

\section{Ethical Concerns}

This paper discusses how different types of language models behave on long-context data. It follows that mistakes in our methodology (both experimental and analytical) could lead to unsupported confidence or skepticism about LLMs. Though neither are unethical, unsupported confidence can be very dangerous. However, given that the overall claim is that LLMs should not be assumed to support context length that extend beyond what they have trained, regardless of their training data, we do not think this paper in itself could be misinterpreted for particularly dangerous outcomes.

As for model choices, we use publicly available models where the license agreements do not restrict what we can say about the model. This should give the reader confidence that our views are unbiased. This is unlike ChatGPT or GPT4, which include an unrestricted indemnity-clause in their license agreement, which could make us financially liable for damages.

\section*{Acknowledgements}

Jerry Huang is supported by a National Science and Engineering Research Council (NSERC) Canada Graduate Scholarship, a Fonds de Recherche du Qu\'{e}bec Nature et technologies (FRQNT) Training Scholarship and a Hydro-Qu\'{e}bec Excellence Scholarship. The experiments were enabled in part by computational resources provided by Calcul Québec (\url{calculquebec.ca}). The author would like to thank Peng Lu and Qiuhao Zeng for useful discussions during the project.

\bibliography{refs}
\appendix
\section{Technical Implementation Details}\label{app:implementation_details}

\subsection{Models Used}

\autoref{tab:models} lists the public models we use for our experimentation.

\begin{table*}[ht!]
    \resizebox{\linewidth}{!}{
        \centering
        \begin{tabular}{l|l|l|c}
        \toprule
        Model & Abbreviation & Public Model Name & HuggingFace Model \\
        \midrule
        \texttt{Mamba2} & \texttt{Mamba2} & \href{https://huggingface.co/state-spaces/mamba2-2.7b}{\tt state-spaces/mamba2-2.7b} & \ding{56} \\
        \texttt{Mamba2Attention} & \texttt{M2A} & \href{https://huggingface.co/state-spaces/mamba2attn-2.7b}{\tt state-spaces/mamba2attn-2.7b} & \ding{56}\\
        \texttt{Transformer++} & \texttt{TPP} & \href{https://huggingface.co/state-spaces/transformerpp-2.7b}{\tt state-spaces/transformerpp-2.7b} & \ding{56} \\
        \texttt{RWKV} & \texttt{RWKV} & \href{https://huggingface.co/RWKV/rwkv-6-world-3b-v2.1}{\tt RWKV/rwkv-6-world-3b-v2.1} & \ding{52} \\
        \texttt{Sheared-LLaMA} & \texttt{SL} & \href{https://huggingface.co/princeton-nlp/Sheared-LLaMA-2.7B}{\tt princeton-nlp/Sheared-LLaMA-2.7B} & \ding{52} \\
        \texttt{Sheared-LLaMA-ShareGPT} & \texttt{SL-IT} & \href{https://huggingface.co/princeton-nlp/Sheared-LLaMA-2.7B-ShareGPT}{\tt princeton-nlp/Sheared-LLaMA-2.7B-ShareGPT} & \ding{52} \\
        \texttt{RecurrentGemma-2B} & \texttt{RG} & \href{https://huggingface.co/google/recurrentgemma-2b}{\tt google/recurrentgemma-2b} & \ding{52} \\
        \texttt{RecurrentGemma-2B-IT} & \texttt{RG-IT} & \href{https://huggingface.co/google/recurrentgemma-2b-it}{\tt google/recurrentgemma-2b-it} & \ding{52} \\
        \bottomrule
        \end{tabular}
    }
    \caption{Models used and public links to their weights.}
    \label{tab:models}
\end{table*}

\subsection{Computing Resources Used}

All experiments were conduced using a single NVIDIA A100 80GB SXM GPU with 6 CPU worker cores. Experiments are run using PyTorch Version 2.2.0 and CUDA 11.8.

\section{The \textsc{Ruler} Benchmark}\label{app:ruler}
To conduct our study, we focus on the \textsc{Ruler} benchmark~\citep{hsieh2024ruler}, which comprises of tasks spanning across four categories: \emph{retrieval}, \emph{multi-hop tracing}, \emph{aggregation}, and \emph{question answering}. We use a publicly available repository\footnote{\url{https://github.com/NVIDIA/RULER}} to generate evaluation examples based on specific input configurations (see \autoref{tab:example-task} for example configurations) that define the length and complexity of each input. In \textsc{RULER}, the task complexity can be thought of as a function of the number of target output tokens and the signal-to-noise ratio in the context. For our experiments, we use the default set of tasks pre-defined by \citet{hsieh2024ruler}.

\begin{table}[ht!]
\centering
\resizebox{\linewidth}{!}{
\begin{tabular}[t]{@{}llp{0.8\linewidth}@{}}
\toprule
\textbf{Task} & \textbf{Configuration} & \textbf{Example} \\
\midrule
\begin{tabular}[t]{@{}l@{}}Single\\NIAH\\(S-NIAH)\end{tabular} & 
\begin{tabular}[t]{@{}l@{}}type\_key = word\\type\_value = number\\type\_haystack = essay\\size\_haystack $\propto$ context length\end{tabular} & 
\begin{tabular}[t]{@{}p{\linewidth}@{}}
\textcolor{lightgray}{(essays) ......} \\
One of the special magic numbers for \textcolor{violet}{long-context} is: \textcolor{orange}{12345}. \textcolor{lightgray}{......}  \\
What is the special magic number for \textcolor{violet}{long-context} mentioned in the provided text?\\
Answer: \textcolor{orange}{12345}
\end{tabular}
\\
\midrule
\begin{tabular}[t]{@{}l@{}}Multi-keys\\NIAH\\(MK-NIAH)\end{tabular}  &
\begin{tabular}[t]{@{}l@{}}num\_keys = 2\\type\_key = word\\type\_value = number\\type\_haystack = essay\\size\_haystack $\propto$ context length\end{tabular} & 
\begin{tabular}[t]{@{}p{\linewidth}@{}}
\textcolor{lightgray}{(essays) ......} \\ 
One of the special magic numbers for \textcolor{violet}{long-context} is: \textcolor{orange}{12345}. \\  
\textcolor{lightgray}{One of the special magic numbers for large-model is: 54321}. \\ 
\textcolor{lightgray}{......}  \\
What is the special magic number for \textcolor{violet}{long-context} mentioned in the provided text?\\
Answer: \textcolor{orange}{12345}
\end{tabular}
\\
\midrule
\begin{tabular}[t]{@{}l@{}}Multi-values\\NIAH\\(MV-NIAH)\end{tabular} &
\begin{tabular}[t]{@{}l@{}}num\_values = 2\\type\_key = word\\type\_value = number\\type\_haystack = essay\\size\_haystack $\propto$ context length\end{tabular} & 
\begin{tabular}[t]{@{}p{\linewidth}@{}}
\textcolor{lightgray}{(essays) ......} \\ 
One of the special magic numbers for \textcolor{violet}{long-context} is: \textcolor{orange}{12345}. \\  
One of the special magic numbers for \textcolor{violet}{long-context} is: \textcolor{orange}{54321}. \\  
\textcolor{lightgray}{......}  \\
What are all the special magic numbers for \textcolor{violet}{long-context} mentioned in the provided text?\\
Answer: \textcolor{orange}{12345}  \textcolor{orange}{54321}
\end{tabular}
\\
\midrule
\begin{tabular}[t]{@{}l@{}}Multi-queries\\NIAH\\(MQ-NIAH)\end{tabular} &
\begin{tabular}[t]{@{}l@{}}num\_queries = 2\\type\_key = word\\type\_value = number\\type\_haystack = essay\\size\_haystack $\propto$ context length\end{tabular} &  
\begin{tabular}[t]{@{}p{\linewidth}@{}}
\textcolor{lightgray}{(essays) ......} \\ 
One of the special magic numbers for \textcolor{violet}{long-context} is: \textcolor{orange}{12345}. \\  
One of the special magic numbers for \textcolor{violet}{large-model} is: \textcolor{orange}{54321}. \\  
\textcolor{lightgray}{......}  \\
What are all the special magic numbers for \textcolor{violet}{long-context} and \textcolor{violet}{large-model} mentioned in the provided text?\\
Answer: \textcolor{orange}{12345}  \textcolor{orange}{54321}
\end{tabular}
\\
\midrule
\begin{tabular}[t]{@{}l@{}}Variable\\Tracking\\(VT)\end{tabular} &
\begin{tabular}[t]{@{}l@{}}num\_chains = 2\\num\_hops = 2\\size\_noises $\propto$ context length\end{tabular} &
\begin{tabular}[t]{@{}p{\linewidth}@{}}
\textcolor{lightgray}{(noises) ......} \\
VAR \textcolor{orange}{X1} = \textcolor{violet}{12345} \textcolor{lightgray}{...... VAR Y1 = 54321 ......}  \\
VAR \textcolor{orange}{X2} = \textcolor{orange}{X1} \textcolor{lightgray}{...... VAR Y2 = Y1 ......} \\
VAR \textcolor{orange}{X3} = \textcolor{orange}{X2} \textcolor{lightgray}{...... VAR Y3 = Y2 ......} \\
Find all variables that are assigned the value \textcolor{violet}{12345}. \\
Answer: \textcolor{orange}{X1 X2 X3}
\end{tabular}
\\
\midrule
\begin{tabular}[t]{@{}l@{}}Common Words\\Extraction\\(CWE)\end{tabular} &
\begin{tabular}[t]{@{}l@{}}freq\_cw = 2, freq\_ucw = 1\\num\_cw = 10\\num\_ucw $\propto$ context length\end{tabular} & 
\begin{tabular}[t]{@{}p{\linewidth}@{}}
\textcolor{orange}{aaa} \textcolor{lightgray}{bbb} \textcolor{orange}{ccc} \textcolor{orange}{aaa} \textcolor{lightgray}{ddd} \textcolor{lightgray}{eee} \textcolor{orange}{ccc} \textcolor{lightgray}{fff} \textcolor{lightgray}{ggg} 
\textcolor{lightgray}{hhh} \textcolor{orange}{iii} \textcolor{orange}{iii} \textcolor{lightgray}{......}\\
What are the 10 most common words in the above list? \\
Answer: \textcolor{orange}{aaa ccc iii ......}
\end{tabular}
\\
\midrule
\begin{tabular}[t]{@{}l@{}}Frequent Words\\Extraction\\(FWE)\end{tabular} &
\begin{tabular}[t]{@{}l@{}}$\alpha$ = 2\\num\_word $\propto$ context length\end{tabular} & 
\begin{tabular}[t]{@{}p{\linewidth}@{}}
\textcolor{orange}{aaa} \textcolor{lightgray}{bbb} \textcolor{orange}{ccc} \textcolor{orange}{aaa} \textcolor{lightgray}{ddd} \textcolor{lightgray}{eee} \textcolor{orange}{ccc} \textcolor{lightgray}{fff} \textcolor{lightgray}{ggg} \textcolor{orange}{aaa} \textcolor{lightgray}{hhh} \textcolor{orange}{aaa} \textcolor{orange}{ccc} \textcolor{orange}{iii} \textcolor{orange}{iii}  \textcolor{lightgray}{......}\\
What are the 3 most frequently appeared words in the above coded text? \\
Answer: \textcolor{orange}{aaa ccc iii}
\end{tabular}
\\
\midrule
\begin{tabular}[t]{@{}l@{}}Question\\Answering\\(QA)\end{tabular} &
\begin{tabular}[t]{@{}l@{}}dataset = SQuAD\\num\_document $\propto$ context length\end{tabular} & 
\begin{tabular}[t]{@{}p{\linewidth}@{}}
\textcolor{lightgray}{Document 1: ...... aaa ......} \\
\textcolor{violet}{Document 2:} \textcolor{lightgray}{......} \textcolor{orange}{bbb} \textcolor{lightgray}{......} \\
\textcolor{lightgray}{Document 3: ...... ccc ......} \\
Question: \textcolor{violet}{question} \\
Answer: \textcolor{orange}{bbb}
\end{tabular}
\\
\bottomrule
\end{tabular}}
\caption{Task examples with flexible configurations in \textsc{Ruler}. 
Different colors highlight \textcolor{violet}{queries}, \textcolor{violet}{keys}, \textcolor{orange}{values}, and \textcolor{lightgray}{distractors} in each example. Examples are retrieved directly from \citet{hsieh2024ruler}.}
\label{tab:example-task}
\end{table}
\subsection{Retrieval: Needle-in-a-haystack (NIAH)} 

\textsc{Ruler} includes multiple retrieval-based tasks, extending the vanilla NIAH test to evaluate models based to four NIAH tasks. The ``needle'' in each of these tasks is a \emph{key-value} pair inserted into the ``haystack'' (long distractor texts). The \emph{query} is located at the end of the sequence and serves as a cue for matching the \emph{keys} in the context and subsequently retrieving the associated \emph{values}. 
\begin{itemize}[leftmargin=0.8em]
\itemsep0em 
\item \textbf{Single NIAH (S-NIAH):} This comprises the standard/vanilla NIAH task where a single ``needle'' needs to be retrieved from the ``haystack''. The \emph{query}/\emph{key}/\emph{value} can take the form of words, numbers (7 digits), or UUIDs (32 digits). The ``haystack'' can be repeated noise sentences or Paul Graham essays~\citep{needle}.

\item \textbf{Multi-keys NIAH (MK-NIAH):} Multiple ``needles'' are inserted into the ``haystack'', and only one of them needs to be retrieved. The additional ``needles'' are hard distractors. The most challenging setting is a version where the ``haystack'' is filled with distractor needles.

\item \textbf{Multi-values NIAH (MV-NIAH):} Multiple ``needles'' sharing the same \emph{key} are inserted into the ``haystack''. All \emph{values} associated with the same \emph{key} need to be retrieved. 

\item \textbf{Multi-queries NIAH (MQ-NIAH):} Multiple ``needles'' are inserted into the ``haystack''. All ``needles'' with distinct keys need to be retrieved. This is the same \emph{multi-query associative recall} task setup used by~\citet{mqar}. Together with MV-NIAH, these two tasks evaluate the retrieval capability without missing any critical information.
\end{itemize}

\subsection{Multi-hop Tracing: Variable Tracking (VT)} 

\emph{Variable tracking} emulates a minimal coreference chain resolution~\citep{ng-2010-supervised} task. This task checks the behavior of tracking relevant co-occurrence patterns and drawing skipped connections within long input. Specifically, a variable $X_1$ is initialized with a value $V$, followed by a linear \emph{chain} of variable name binding statements (e.g., $X_2 = X_1, X_3 = X_2, ...$), which are inserted at various positions of the input. The objective is to return all variable names pointing to $V$. The task complexity can be increased by adding more hops (i.e., the times of name binding) or more chains. 

\subsection{Aggregation: Common Words (CWE) and Frequent Words Extraction (FWE)}
In the common word extraction task (CWE), words are sampled from discrete uniform distributions, with the number of common words fixed while the number of uncommon words increases with the sequence length. In the frequent words extraction task (FWE), words are sampled from a Zeta distribution. A model needs to return the top-$K$ frequent words in the context. In CWE, $K$ is equivalent to the number of common words. In FWE, $K$ is set to 3, as \citet{hsieh2024ruler} observe that increasing $K$ leads to poor performance even at small context sizes for most models. The task complexity can be adjusted by varying the number of common words or the parameter of the Zeta distribution. 

\subsection{Question Answering (QA)}
The majority of existing QA datasets~\citep{squad, hotpotqa, musique} are designed to answer questions based on short passages. These datasets can be extended to simulate long-context input by adding distracting information. In this task category, we insert the golden paragraphs (i.e., the paragraphs that contain answers) into paragraphs randomly sampled from the same dataset. This category is a real-world adaptation~\citep{sled} of NIAH, where the question serves as the query, the golden paragraphs are the ``needles'', and the distracting paragraphs form the ``haystack''.

\section{\textsc{Ruler} Task Results}\label{app:ruler_tasks}

\begin{table}[ht!]
    \centering
    \resizebox{\linewidth}{!}{
        \begin{tabular}{l|ccccc|c}
        \toprule
        Length & \texttt{1K} & \texttt{2K} & \texttt{4K} & \texttt{8K} & \texttt{16K} & Average \\
        \midrule
        \texttt{Mamba2} & 66.8 & 71.6 & 60.0 & 62.4 & 0.0 & 52.16 \\
        \texttt{M2A} & 58.0 & 36.4 & 43.2 & 18.4 & 0.0 & 31.2 \\
        \texttt{TPP} & 40.4 & 24.8 & 0.0 & 0.0 & 0.0 & 13.04 \\
        \midrule
        \texttt{RG} & 100.0 & 100.0 & 52.0 & 24.8 & 10.0 & 57.36 \\
        \texttt{SL} & 100.0 & 100.0 & 100.0 & 0.0 & 0.0 & 60.0 \\
        \texttt{RWKV} & 100.0 & 100.0 & 100.0 & 100.0 & 54.4 & 90.88 \\
        \midrule
        \texttt{RG-IT} & 100.0 & 100.0 & 51.6 & 28.8 & 16.4 & 59.36 \\
        \texttt{SL-IT} & 100.0 & 100.0 & 100.0 & 0.0 & 0.0 & 60.0 \\
        \bottomrule
        \end{tabular}
    }
    \caption{Results on \texttt{niah\_single\_1} task of \textsc{Ruler}.}
    \label{tab:ruler_niah_single_1}
\end{table}

\begin{table}[ht!]
        \centering
    \resizebox{\linewidth}{!}{
        \begin{tabular}{l|ccccc|c}
        \toprule
        Length & \texttt{1K} & \texttt{2K} & \texttt{4K} & \texttt{8K} & \texttt{16K} & Average \\
        \midrule
        \texttt{Mamba2} & 62.4 & 60.4 & 0.0 & 0.0 & 0.0 & 24.56 \\
        \texttt{M2A} & 33.2 & 34.8 & 9.6 & 4.8 & 0.0 & 16.48 \\
        \texttt{TPP} & 50.8 & 48.0 & 0.0 & 0.0 & 0.0 & 19.76 \\
        \midrule
        \texttt{RG} & 100.0 & 100.0 & 36.4 & 16.8 & 2.8 & 51.2 \\
        \texttt{SL} & 99.6 & 99.6 & 100.0 & 0.0 & 0.0 & 59.84 \\
        \texttt{RWKV} & 100.0 & 100.0 & 53.6 & 30.4 & 9.6 & 58.72 \\
        \midrule
        \texttt{RG-IT} & 100.0 & 100.0 & 55.2 & 24.4 & 12.8 & 58.48 \\
        \texttt{SL-IT} & 100.0 & 100.0 & 100.0 & 0.0 & 0.0 & 60.0 \\
        \bottomrule
        \end{tabular}
    }
    \caption{Results on \texttt{niah\_single\_2} task of \textsc{Ruler}.}
    \label{tab:ruler_niah_single_2}
\end{table}

\begin{table}[ht!]
        \centering
    \resizebox{\linewidth}{!}{
        \begin{tabular}{l|ccccc|c}
        \toprule
        Length & \texttt{1K} & \texttt{2K} & \texttt{4K} & \texttt{8K} & \texttt{16K} & Average \\
        \midrule
\texttt{Mamba2} & 52.0 & 61.6 & 0.0 & 0.0 & 0.0 & 22.72 \\
\texttt{M2A} & 38.8 & 32.4 & 2.8 & 6.4 & 0.0 & 16.08 \\
\texttt{TPP} & 64.4 & 53.2 & 0.0 & 0.0 & 0.0 & 23.52 \\
\midrule
\texttt{RG} & 100.0 & 100.0 & 39.2 & 16.8 & 8.4 & 52.88 \\
\texttt{SL} & 100.0 & 100.0 & 96.4 & 0.0 & 0.0 & 59.28 \\
\texttt{RWKV} & 99.2 & 96.4 & 15.2 & 19.6 & 4.4 & 46.96 \\
\midrule
\texttt{RG-IT} & 100.0 & 100.0 & 53.6 & 24.0 & 13.6 & 58.24 \\
\texttt{SL-IT} & 100.0 & 99.6 & 99.6 & 0.0 & 0.0 & 59.84 \\
        \bottomrule
        \end{tabular}
    }
    \caption{Results on \texttt{niah\_single\_3} task of \textsc{Ruler}.}
    \label{tab:ruler_niah_single_3}
\end{table}

\begin{table}[ht!]
        \centering
    \resizebox{\linewidth}{!}{
        \begin{tabular}{l|ccccc|c}
        \toprule
        Length & \texttt{1K} & \texttt{2K} & \texttt{4K} & \texttt{8K} & \texttt{16K} & Average \\
        \midrule
\texttt{Mamba2} & 25.6 & 23.6 & 0.0 & 0.0 & 0.0 & 9.84 \\
\texttt{M2A} & 21.2 & 16.4 & 5.2 & 1.2 & 0.0 & 8.8 \\
\texttt{TPP} & 50.0 & 34.4 & 0.0 & 0.0 & 0.0 & 16.88 \\
\midrule
\texttt{RG} & 98.8 & 98.8 & 23.2 & 15.6 & 4.4 & 48.16 \\
\texttt{SL} & 99.2 & 100.0 & 94.0 & 0.0 & 0.0 & 58.64 \\
\texttt{RWKV} & 81.6 & 64.0 & 30.4 & 18.0 & 11.2 & 41.04 \\
\midrule
\texttt{RG-IT} & 99.2 & 100.0 & 36.8 & 17.6 & 11.2 & 52.96 \\
\texttt{SL-IT} & 99.6 & 99.2 & 98.0 & 0.0 & 0.0 & 59.36 \\
        \bottomrule
        \end{tabular}
    }
    \caption{Results on \texttt{niah\_multikey\_1} task of \textsc{Ruler}.}
    \label{tab:ruler_niah_multikey_1}
\end{table}

\begin{table}[ht!]
        \centering
    \resizebox{\linewidth}{!}{
        \begin{tabular}{l|ccccc|c}
        \toprule
        Length & \texttt{1K} & \texttt{2K} & \texttt{4K} & \texttt{8K} & \texttt{16K} & Average \\
        \midrule
\texttt{Mamba2} & 4.8 & 2.0 & 0.0 & 0.0 & 0.0 & 1.36 \\
\texttt{M2A} & 17.2 & 7.6 & 0.4 & 0.0 & 0.0 & 5.04 \\
\texttt{TPP} & 60.0 & 36.4 & 0.0 & 0.0 & 0.0 & 19.28 \\
\midrule
\texttt{RG} & 98.0 & 94.8 & 8.4 & 2.4 & 1.6 & 41.04 \\
\texttt{SL} & 95.2 & 86.8 & 53.6 & 0.0 & 0.0 & 47.12 \\
\texttt{RWKV} & 20.4 & 4.0 & 0.8 & 0.4 & 0.0 & 5.12 \\
\midrule
\texttt{RG-IT} & 100.0 & 98.0 & 43.6 & 27.2 & 9.6 & 55.68 \\
\texttt{SL-IT} & 97.6 & 96.0 & 78.8 & 0.0 & 0.0 & 54.48 \\
        \bottomrule
        \end{tabular}
    }
    \caption{Results on \texttt{niah\_multikey\_2} task of \textsc{Ruler}.}
    \label{tab:ruler_niah_multikey_2}
\end{table}

\begin{table}[ht!]
        \centering
    \resizebox{\linewidth}{!}{
        \begin{tabular}{l|ccccc|c}
        \toprule
        Length & \texttt{1K} & \texttt{2K} & \texttt{4K} & \texttt{8K} & \texttt{16K} & Average \\
        \midrule
\texttt{Mamba2} & 14.4 & 2.4 & 0.0 & 0.0 & 0.0 & 3.36 \\
\texttt{M2A} & 17.6 & 12.4 & 0.0 & 0.0 & 0.0 & 6.0 \\
\texttt{TPP} & 61.2 & 56.4 & 0.0 & 0.0 & 0.0 & 23.52 \\
\midrule
\texttt{RG} & 74.8 & 58.8 & 7.2 & 2.8 & 1.6 & 29.04 \\
\texttt{SL} & 96.4 & 46.4 & 38.8 & 0.0 & 0.0 & 36.32 \\
\texttt{RWKV} & 14.8 & 1.6 & 0.4 & 0.0 & 0.0 & 3.36 \\
\midrule
\texttt{RG-IT} & 88.0 & 92.0 & 16.0 & 14.0 & 1.6 & 42.32 \\
\texttt{SL-IT} & 85.6 & 63.2 & 59.2 & 0.0 & 0.0 & 41.6 \\
        \bottomrule
        \end{tabular}
    }
    \caption{Results on \texttt{niah\_multikey\_3} task of \textsc{Ruler}.}
    \label{tab:ruler_niah_multikey_3}
\end{table}

\begin{table}[ht!]
        \centering
    \resizebox{\linewidth}{!}{
        \begin{tabular}{l|ccccc|c}
        \toprule
        Length & \texttt{1K} & \texttt{2K} & \texttt{4K} & \texttt{8K} & \texttt{16K} & Average \\
        \midrule
\texttt{Mamba2} & 34.9 & 26.6 & 0.0 & 0.0 & 0.0 & 12.3 \\
\texttt{M2A} & 48.8 & 33.5 & 1.3 & 0.1 & 0.0 & 16.74 \\
\texttt{TPP} & 42.3 & 31.1 & 0.0 & 0.0 & 0.0 & 14.68 \\
\midrule
\texttt{RG} & 97.4 & 95.1 & 14.7 & 3.3 & 3.0 & 42.7 \\
\texttt{SL} & 100.0 & 82.5 & 44.0 & 0.0 & 0.0 & 45.3 \\
\texttt{RWKV} & 96.5 & 87.0 & 57.2 & 10.8 & 5.2 & 51.34 \\
\midrule
\texttt{RG-IT} & 96.7 & 87.6 & 41.8 & 22.0 & 11.3 & 51.88 \\
\texttt{SL-IT} & 100.0 & 87.5 & 77.2 & 0.0 & 0.0 & 52.94 \\
        \bottomrule
        \end{tabular}
    }
    \caption{Results on \texttt{niah\_multivalue} task of \textsc{Ruler}.}
    \label{tab:ruler_niah_multivalue}
\end{table}

\begin{table}[ht!]
        \centering
    \resizebox{\linewidth}{!}{
        \begin{tabular}{l|ccccc|c}
        \toprule
        Length & \texttt{1K} & \texttt{2K} & \texttt{4K} & \texttt{8K} & \texttt{16K} & Average \\
        \midrule
\texttt{Mamba2} & 39.1 & 39.2 & 0.0 & 0.0 & 0.0 & 15.66 \\
\texttt{M2A} & 54.4 & 37.5 & 1.6 & 0.0 & 0.0 & 18.7 \\
\texttt{TPP} & 44.4 & 34.8 & 0.0 & 0.0 & 0.0 & 15.84 \\
\midrule
\texttt{RG} & 99.5 & 99.7 & 4.7 & 2.8 & 2.8 & 41.9 \\
\texttt{SL} & 98.8 & 80.8 & 45.6 & 0.0 & 0.0 & 45.04 \\
\texttt{RWKV} & 94.3 & 80.7 & 38.4 & 9.3 & 2.4 & 45.02 \\
\midrule
\texttt{RG-IT} & 97.8 & 97.9 & 48.5 & 21.1 & 11.4 & 55.34 \\
\texttt{SL-IT} & 98.4 & 94.7 & 85.9 & 0.0 & 0.0 & 55.8 \\
        \bottomrule
        \end{tabular}
    }
    \caption{Results on \texttt{niah\_multiquery} task of \textsc{Ruler}.}
    \label{tab:ruler_niah_multiquery}
\end{table}

\begin{table}[ht!]
        \centering
    \resizebox{\linewidth}{!}{
        \begin{tabular}{l|ccccc|c}
        \toprule
        Length & \texttt{1K} & \texttt{2K} & \texttt{4K} & \texttt{8K} & \texttt{16K} & Average \\
        \midrule
\texttt{Mamba2} & 69.12 & 36.64 & 35.2 & 20.72 & 0.0 & 32.34 \\
\texttt{M2A} & 78.24 & 56.88 & 9.6 & 1.76 & 0.56 & 29.41 \\
\texttt{TPP} & 40.88 & 21.12 & 0.0 & 0.0 & 0.0 & 12.4 \\
\midrule
\texttt{RG} & 98.0 & 75.52 & 0.0 & 0.0 & 0.0 & 34.7 \\
\texttt{SL} & 98.16 & 81.68 & 19.36 & 0.0 & 0.0 & 39.84 \\
\texttt{RWKV} & 68.56 & 47.76 & 20.08 & 6.88 & 10.95 & 30.85 \\
\midrule
\texttt{RG-IT} & 84.24 & 79.36 & 50.4 & 31.76 & 19.92 & 53.14 \\
\texttt{SL-IT} & 93.68 & 76.88 & 42.32 & 0.0 & 0.0 & 42.58 \\
        \bottomrule
        \end{tabular}
    }
    \caption{Results on \texttt{vt} task of \textsc{Ruler}.}
    \label{tab:ruler_vt}
\end{table}

\begin{table}[ht!]
        \centering
    \resizebox{\linewidth}{!}{
        \begin{tabular}{l|ccccc|c}
        \toprule
        Length & \texttt{1K} & \texttt{2K} & \texttt{4K} & \texttt{8K} & \texttt{16K} & Average \\
        \midrule
\texttt{Mamba2} & 28.52 & 14.72 & 4.08 & 0.16 & 0.12 & 9.52 \\
\texttt{M2A} & 26.48 & 15.24 & 3.04 & 5.92 & 0.8 & 10.3 \\
\texttt{TPP} & 30.32 & 17.8 & 0.64 & 0.0 & 0.04 & 9.76 \\
\midrule
\texttt{RG} & 48.6 & 21.32 & 42.88 & 17.24 & 4.24 & 26.86 \\
\texttt{SL} & 71.2 & 25.32 & 55.24 & 0.0 & 0.04 & 30.36 \\
\texttt{RWKV} & 57.08 & 3.24 & 45.0 & 14.84 & 1.92 & 24.42 \\
\midrule
\texttt{RG-IT} & 55.4 & 4.56 & 17.4 & 3.24 & 0.2 & 16.16 \\
\texttt{SL-IT} & 78.96 & 18.64 & 57.2 & 0.0 & 0.0 & 30.96 \\
        \bottomrule
        \end{tabular}
    }
    \caption{Results on \texttt{cwe} task of \textsc{Ruler}.}
    \label{tab:ruler_cwe}
\end{table}

\begin{table}[ht!]
        \centering
    \resizebox{\linewidth}{!}{
        \begin{tabular}{l|ccccc|c}
        \toprule
        Length & \texttt{1K} & \texttt{2K} & \texttt{4K} & \texttt{8K} & \texttt{16K} & Average \\
        \midrule
\texttt{Mamba2} & 57.87 & 44.67 & 40.67 & 0.53 & 0.0 & 28.75 \\
\texttt{M2A} & 59.73 & 53.73 & 58.0 & 52.0 & 39.6 & 52.61 \\
\texttt{TPP} & 59.6 & 56.4 & 0.13 & 0.0 & 0.0 & 23.23 \\
\midrule
\texttt{RG} & 56.0 & 53.87 & 7.6 & 15.6 & 17.33 & 30.08 \\
\texttt{SL} & 72.0 & 38.67 & 45.07 & 0.0 & 0.0 & 31.15 \\
\texttt{RWKV} & 74.67 & 67.47 & 68.0 & 56.67 & 43.42 & 62.05 \\
\midrule
\texttt{RG-IT} & 80.8 & 67.87 & 69.73 & 64.8 & 50.67 & 66.77 \\
\texttt{SL-IT} & 78.67 & 78.27 & 73.07 & 0.0 & 0.0 & 46.0 \\
        \bottomrule
        \end{tabular}
    }
    \caption{Results on \texttt{fwe} task of \textsc{Ruler}.}
    \label{tab:ruler_fwe}
\end{table}

\begin{table}[ht!]
        \centering
    \resizebox{\linewidth}{!}{
        \begin{tabular}{l|ccccc|c}
        \toprule
        Length & \texttt{1K} & \texttt{2K} & \texttt{4K} & \texttt{8K} & \texttt{16K} & Average \\
        \midrule
\texttt{Mamba2} & 25.2 & 24.4 & 18.4 & 0.0 & 0.4 & 13.68 \\
\texttt{M2A} & 33.6 & 35.6 & 18.0 & 6.8 & 3.6 & 19.52 \\
\texttt{TPP} & 37.2 & 36.4 & 2.8 & 0.8 & 0.4 & 15.52 \\
\midrule
\texttt{RG} & 26.8 & 15.6 & 31.2 & 6.8 & 8.8 & 17.84 \\
\texttt{SL} & 41.6 & 37.2 & 37.2 & 0.0 & 0.0 & 23.2 \\
\texttt{RWKV} & 46.4 & 35.6 & 30.8 & 21.2 & 18.4 & 30.48 \\
\midrule
\texttt{RG-IT} & 74.0 & 66.8 & 58.4 & 10.0 & 9.6 & 43.76 \\
\texttt{SL-IT} & 54.4 & 56.4 & 55.6 & 0.0 & 0.0 & 33.28 \\
        \bottomrule
        \end{tabular}
    }
    \caption{Results on \texttt{qa\_1} task of \textsc{Ruler}.}
    \label{tab:ruler_qa_1}
\end{table}

\begin{table}[ht!]
        \centering
    \resizebox{\linewidth}{!}{
        \begin{tabular}{l|ccccc|c}
        \toprule
        Length & \texttt{1K} & \texttt{2K} & \texttt{4K} & \texttt{8K} & \texttt{16K} & Average \\
        \midrule
\texttt{Mamba2} & 20.0 & 20.0 & 10.4 & 0.8 & 0.8 & 10.4 \\
\texttt{M2A} & 21.6 & 23.2 & 14.8 & 4.0 & 0.8 & 12.88 \\
\texttt{TPP} & 24.4 & 26.8 & 0.4 & 0.0 & 0.0 & 10.32 \\
\midrule
\texttt{RG} & 26.8 & 18.8 & 24.4 & 20.8 & 16.8 & 21.52 \\
\texttt{SL} & 24.8 & 29.6 & 29.6 & 0.0 & 0.0 & 16.8 \\
\texttt{RWKV} & 31.6 & 30.8 & 27.2 & 20.4 & 17.6 & 25.52 \\
\midrule
\texttt{RG-IT} & 37.2 & 38.8 & 33.2 & 25.6 & 16.0 & 30.16 \\
\texttt{SL-IT} & 34.0 & 37.6 & 38.4 & 0.0 & 0.0 & 22.0 \\
        \bottomrule
        \end{tabular}
    }
    \caption{Results on \texttt{qa\_2} task of \textsc{Ruler}.}
    \label{tab:ruler_qa_2}
\end{table}


\end{document}